\pdfoutput=1

\documentclass[11pt]{article}

\usepackage[]{acl}

\usepackage{times}
\usepackage{latexsym}

\usepackage[T1]{fontenc}

\usepackage[utf8]{inputenc}
\usepackage{hyperref}       
\usepackage{url}            
\usepackage{booktabs}       
\usepackage{amsfonts}       
\usepackage{nicefrac}       
\usepackage{microtype}      
\usepackage{xcolor}         
\usepackage{subfig}
\usepackage{graphicx}
\usepackage{stfloats}
\usepackage{bbding}
\usepackage{inconsolata}
\usepackage{amsmath}
\usepackage{amsfonts} 
\usepackage{multirow}
\usepackage{algorithm}
\usepackage{algorithmic}
\usepackage{geometry}
\usepackage{mathrsfs}
\usepackage{enumerate}
\usepackage{longtable}
\usepackage{amsthm}
\usepackage{wrapfig}
\usepackage{colortbl}
\usepackage{caption}
\usepackage{tabularx} 
\usepackage{ragged2e} 
\usepackage{booktabs} 
\newcolumntype{L}{>{\RaggedRight\hangafter=1\hangindent=0em}X}
\usepackage{microtype}
\definecolor{mydarkgreen}{RGB}{0,128,0}
\newcommand*\samethanks[1][\value{footnote}]{\footnotemark[#1]}
\newcommand{\removelatexerror}{\let\@latex@error\@gobble}
\title{Progressive Knowledge Graph Completion}

\author{
  Jiayi Li$^{12}$\thanks{~~Equal contribution.},~
  Ruilin Luo$^{1}$\samethanks,~
  Jiaqi Sun$^{3}$,~ 
  Jing Xiao$^{4}$,~
  Yujiu Yang$^{1}$\thanks{~{~Corresponding author: yang.yujiu@sz.tsinghua.edu.cn}} 
  \\$^{1}$Tsinghua University
  \\$^{2}$Baidu Inc.
  \\$^{3}$Carnegie Mellon University
  \\$^{4}$Ping An Technology (Shenzhen) Co., Ltd.
  \\ \texttt{\{lijy20,lrl23\}@mails.tsinghua.edu.cn}
}
\begin{document}
\maketitle
\begin{abstract}
Knowledge Graph Completion (KGC) has emerged as a promising solution to address the issue of incompleteness within Knowledge Graphs (KGs). Traditional KGC research primarily centers on triple classification and link prediction. Nevertheless, we contend that these tasks do not align well with real-world scenarios and merely serve as surrogate benchmarks. In this paper, we investigate three crucial processes relevant to real-world construction scenarios: (a) the \textit{verification process}, which arises from the necessity and limitations of human verifiers; (b) the \textit{mining process}, which identifies the most promising candidates for verification; and (c) the \textit{training process}, which harnesses verified data for subsequent utilization; in order to achieve a transition toward more realistic challenges. By integrating these three processes, we introduce the Progressive Knowledge Graph Completion (PKGC) task, which simulates the gradual completion of KGs in real-world scenarios. Furthermore, to expedite PKGC processing, we propose two acceleration modules: Optimized Top-$k$ algorithm and Semantic Validity Filter. These modules significantly enhance the efficiency of the mining procedure. Our experiments demonstrate that performance in link prediction does not accurately reflect performance in PKGC. A more in-depth analysis reveals the key factors influencing the results and provides potential directions for future research. Codes are available at \href{https://github.com/hyleepp/Continue-KGC}{https://github.com/hyleepp/Continue-KGC}. 
\end{abstract}

\section{Introduction}
{Knowledge} Graphs (KGs) have wide-ranging applications across diverse domains, including question answering~\cite{QA}, information extraction~\cite{acquisition}, and recommender systems~\cite{recommend}. Nevertheless, KGs frequently grapple with incompleteness, resulting in the absence of critical factual links~\cite{KGCompleteness}. Consequently, Knowledge Graph Completion (KGC) assumes a pivotal role in automating the enhancement of KGs~\cite{sun2019re}.

In the past, tasks such as link prediction and triple classification required predicting the tail entity of a query and judging the correctness of the proposed triples, respectively. However, where do these effective queries come from in real-world scenarios? No previous work has explored this issue. Furthermore, the performance of models in these tasks falls short of the high accuracy requirements of KG, as confirmed by prominent companies like Google~\cite{llmaskg}. Therefore, we advocate for simulating a more realistic scenario in KGC. To meet the requirements for stringent knowledge precision, it is necessary to incorporate a \textit{verification process} to simulate human-machine collaboration. Additionally, human verifiers face inherent limitations in their daily data processing capacity, underscoring the need for a \textit{mining process}. Within this process, KGC models curate a specific quota of the most promising facts. These freshly acquired facts, in turn, facilitate the iterative refinement of KGC models through a \textit{training process}. Finally, we iteratively implement these three processes to form a progressive completion.

\begin{figure*}[!t]
    \centering
    \includegraphics[width=0.9\linewidth]{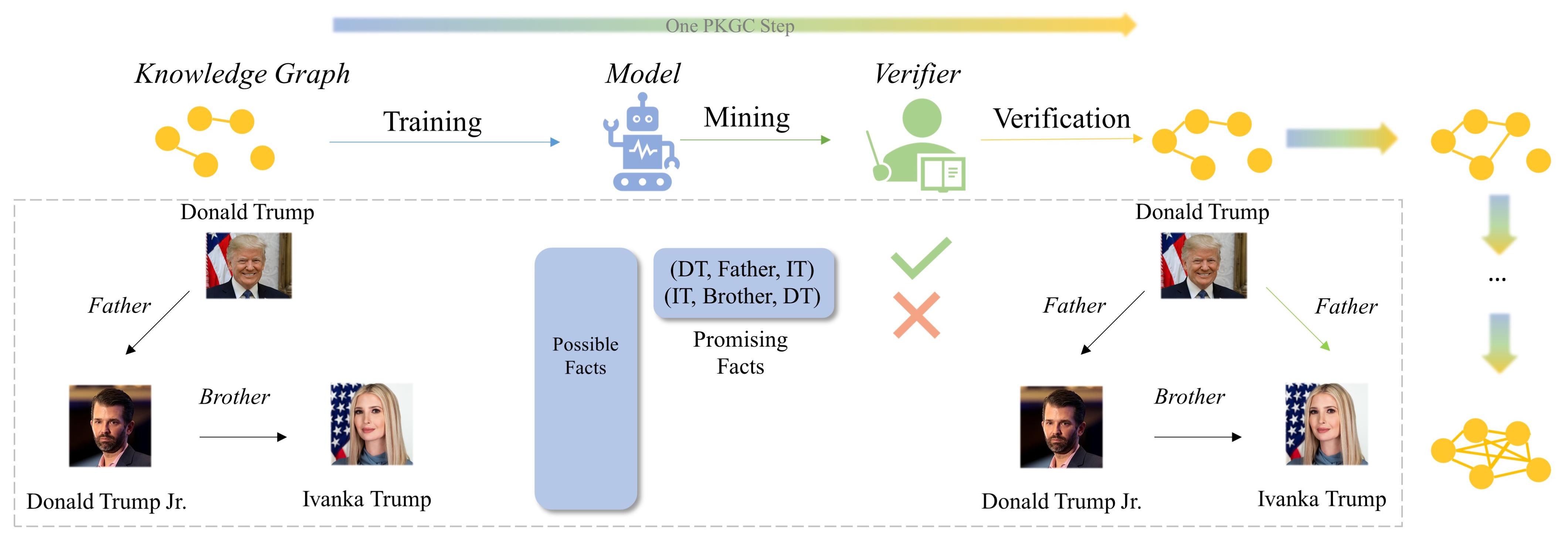}
    \caption{PKGC consists of training, mining and verification procedures. The knowledge proposed by the KGE model will be added to the knowledge base after verification. }
    \label{fig:introduction}
\end{figure*}

Drawing from the above reasons, we introduce the Progressive Knowledge Graph Completion (PKGC) task. PKGC emulates the gradual process of KG completion, as depicted in Figure \ref{fig:introduction}. It commences by training a model using a KG and subsequently invokes the model to identify the most promising facts. Thereafter, a limited verifier selects the true facts, integrating them into the KG.Furthermore, we have made substantial strides in expediting the mining process, particularly vital given the exponential growth in potential facts as KGs expand in size. To address this, we introduce two modules: Optimized Top-$k$ and Semantic Validity Filter (SVF), both of which bring about a significant reduction in both space and time complexity. The former implements root filtering within a heap structure and incorporates a batch warm-up strategy, while the latter capitalizes on the semantic properties inherent to KGs.

In the realm of PKGC, we propose two novel metrics specifically tailored to provide a more realistic evaluation of the model's performance. It becomes apparent that relying on metrics derived from previous Surrogate Knowledge Graph Completion~(SKGC) tasks inadequately guides the model's performance in PKGC. Therefore, they are not suitable for selecting models in real scenarios. In light of this observation, we initiate a discussion to delve into the underlying factors influencing model performance in PKGC. We also explore opportunities to leverage verified knowledge in PKGC, as opposed to SKGC, which models entities and relations in a one-off manner. We leverage two simple incremental learning approaches and throw light on them for future research.

\section{Related Work}

\paragraph{Knowledge Graph Embedding} KGC encompasses a diverse array of methods~\cite{survey1}, including embedding-based~\cite{bordes2013translating, transh, sun2018rotate, trans}, rule-based~\cite{amie, rlvlr}, and reinforcement learning-based~\cite{deeppath, multi-hop}. Given the prevalence of the embedding-based approach, often referred to as Knowledge Graph Embedding (KGE), this article primarily centers its focus on this methodology. Notably, KGE methods have consistently dominated the leaderboards of competitions like ogb-wiki2~\cite{ogb}, a prominent fixture in the field of KGC. To streamline the scope, this paper confines its inquiry to the realm of structural information, excluding description-based approaches~\cite{entity-descrip, text-join-embedding, text, kepler, kg-bert} within KGE. 

The pioneering distance-based model in KGE, TransE~\cite{bordes2013translating}, conceptualizes each relation as a translation operation. TransR~\cite{transr} and TransH~\cite{transh} employ projections to handle complex relations, while MuRP~\cite{murp} extends the modeling space to hyperbolic geometry to capture hierarchical structures within KGs. In addition to translation, RotatE~\cite{sun2018rotate} introduces rotation as a means to represent relations. RotH~\cite{rote} further amalgamates these two operations into hyperbolic space. RESCAL~\cite{rescal} stands as the inaugural bilinear-based model, while CP~\cite{cp} simplifies relation matrices to diagonal representations. Deep learning-based models harness convolutional neural networks~\cite{conve, convkb}, Transformers~\cite{hitter, transformer}, and graph neural networks~\cite{rgcn, gaats, ragat, kracl} as encoders, synergizing them with the aforementioned decoder models.

\section{Methodology}
This section begins with an in-depth exploration of the progressive completion task, further detailed in Section \ref{sec:progress completion}. Following the introduction of our proposed task formulation, we present two acceleration techniques in Section \ref{sec:acceleration for knowledge mining}: the optimized top-$k$ filtering algorithm and the semantic validity filtering module, both of which render PKGC feasible.
\subsection{Progressive Knowledge Graph Completion}
\label{sec:progress completion}
To commence, PKGC initiates by partitioning a KG, denoted as $\mathcal{F}$, into two components: the known portion, refered to as $\mathcal{F}_{known}$, and the unknown portion, labeled $\mathcal{F}_{un}$, based on a predefined ratio $\rho$. This procedure is in alignment with SKGC practices. Distinguishing itself from SKGC, PKGC systematically advances by expanding the known segment, $\mathcal{F}_{known}$, through a sequence of incremental verifications executed by a verifier denoted as $\psi$, with the capability to authenticate $n_c$ candidate facts during each iteration. This process adheres to a cyclic routine that encompasses training, mining, and verification procedures, as elaborated in Algorithm \ref{alg:pkgc}.~(Detailed comparison is in Appendix~\ref{sec:comparison to surrogate completion})
\begin{figure}[H]
    \vspace{-3mm}
  \renewcommand{\algorithmicrequire}{\textbf{Input:}}
  \renewcommand{\algorithmicensure}{\textbf{Output:}}
  \begin{algorithm}[H]
    \caption{Progressive Knowledge Graph Completion}\label{alg:pkgc}
    \begin{algorithmic}[1]
      \REQUIRE KG $\mathcal{F}$, ratio $\rho$, verifier $\psi$, maximum step $n_s$
      \ENSURE Updated KG $\mathcal{F}_{known}$, metrics $\mathcal{M}$
      \STATE \textbf{Initialization:} $s(\cdot)\gets PretrainModel$, $\mathcal{F}_{known}$,$\mathcal{F}_{un}\gets SplitKG(\rho)$,\\ $i \gets 0$, $\mathcal{F}_{visited} \gets \mathcal{F}_{known}$
      \FOR{$i = 1, 2,\dots, n_s$}
        \IF{update condition}
        \STATE $s(\cdot) \gets UpdateModel$
        \ENDIF

        \STATE $\mathcal{F}_{c} \gets KnowledgeMining$ 
        \STATE $\mathcal{F}_{new} \gets Verification$ 
        \STATE $\mathcal{F}_{visited} \gets \mathcal{F}_{visited} \cup \mathcal{F}_c$ 
        \STATE $\mathcal{F}_{known} \gets \mathcal{F}_{known} \cup \mathcal{F}_{new}$ 
      \ENDFOR
      \STATE $M \gets CalculateMetrics$
      
    \end{algorithmic}
  \end{algorithm}
\end{figure}

Within the training phase, we train a model denoted as $s(\cdot)$ using $\mathcal{F}_{known}$, mirroring the methodology employed in Surrogate Knowledge Graph Completion (SKGC). Given that the Knowledge Graph undergoes alterations following each step, $s(\cdot)$ can be subject to a range of update techniques to adapt to these modifications.

The mining phase entails $s(\cdot)$ generating a set of $n_c$ candidate facts, $\mathcal{F}_c$, derived from the entire spectrum of conceivable facts, excluding those present in $\mathcal{F}_{known}$ or validated within $\mathcal{F}_{visited}$.

In the verification phase, the verifier $\psi$ systematically scrutinizes the authenticity of the candidate facts within $\mathcal{F}_c$ and subsequently incorporates the verified facts into the established KG, $\mathcal{F}_{known}$. In the context of our experiments, this process is emulated by verifying the existence of facts within $\mathcal{F}_{un}$.

To assess the performance of PKGC, we establish a predefined maximum number of steps, denoted as $n_s$, and subsequently evaluate the metrics detailed in Section \ref{sec:experiment setting} upon the completion of these steps.

\subsection{Acceleration for Mining Process}
\label{sec:acceleration for knowledge mining}

Given that the mining process can be fundamentally likened to a top-k procedure, executing it without optimization proves to be exceedingly impractical due to the sheer volume of potential facts involved.

To illustrate the magnitude of this issue, consider a KG housing $|\mathcal{E}|$ entities and $|\mathcal{R}|$ relations. Such a KG spawns an astronomical $|\mathcal{E}|^2|\mathcal{R}|$ potential facts, rendering the storage and effective sorting of these facts unviable through conventional means. In response to this formidable challenge, we introduce two purpose-built modules: Optimized Top-$k$ and SVF. These modules are meticulously crafted to achieve substantial reductions in the time and space resources necessary for mining while preserving efficiency and effectiveness.

\subsubsection{Optimized Top-$k$ algorithm}
\label{sec:optimized top-k}

In order to obviate the necessity of retaining scores for every conceivable triple, we have elected to implement a heap structure, which entails only a modest additional space of $O(k)$. Nevertheless, the unrefined heap-based top-$k$ algorithm endeavors to establish a min-heap and proceeds to insert each potential triple into the heap contingent upon its respective score. Notably, despite the algorithm's time complexity of $O(n\log(k))$~\cite{introduction_algorithms}, its practicality is hampered by the vast scale of $|\mathcal{E}|^2|\mathcal{R}|$.

To address this issue, we introduce an optimized top-$k$ algorithm that builds upon the fundamental approach. This algorithm comprises two intricately interconnected components: the root filter and the warm-up.
    
\paragraph{Root Filter} In light of the impracticality of acquiring scores for all conceivable facts in a single sweep, we adopt a strategy of segregating them into a sequence of incremental batches. This sequential approach offers us a unique opportunity to refine the filtration process within each batch by leveraging insights gleaned from previous batches.

\begin{figure*}[!t]
	\centering
    \vspace{-10pt}
	\subfloat[Root filter.]{\includegraphics[width=3.0in]{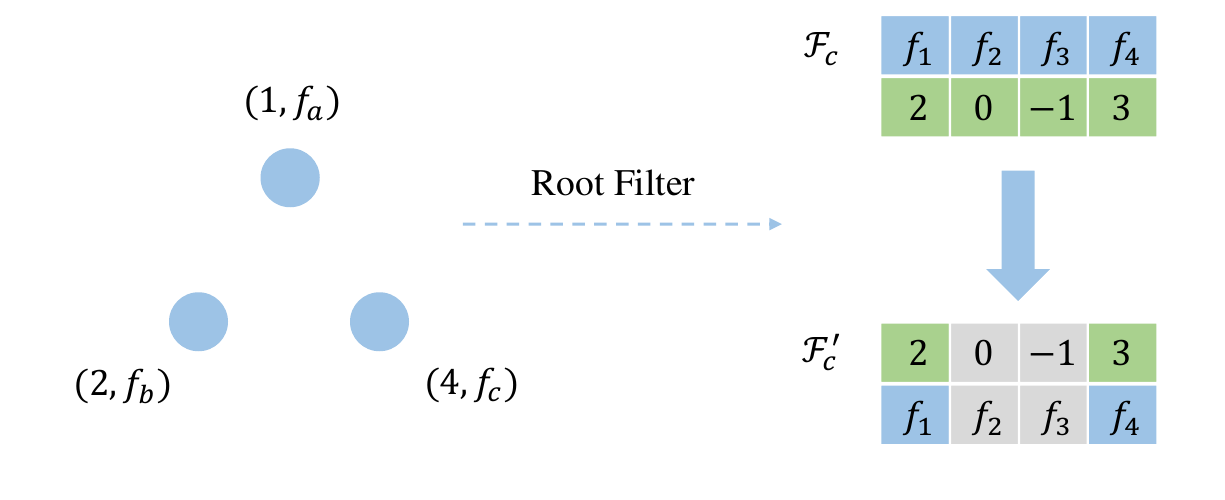}%
		\label{fig:method:root filter}}
	\hfil
	\subfloat[Batch warm-up.]{\includegraphics[width=3.0in]{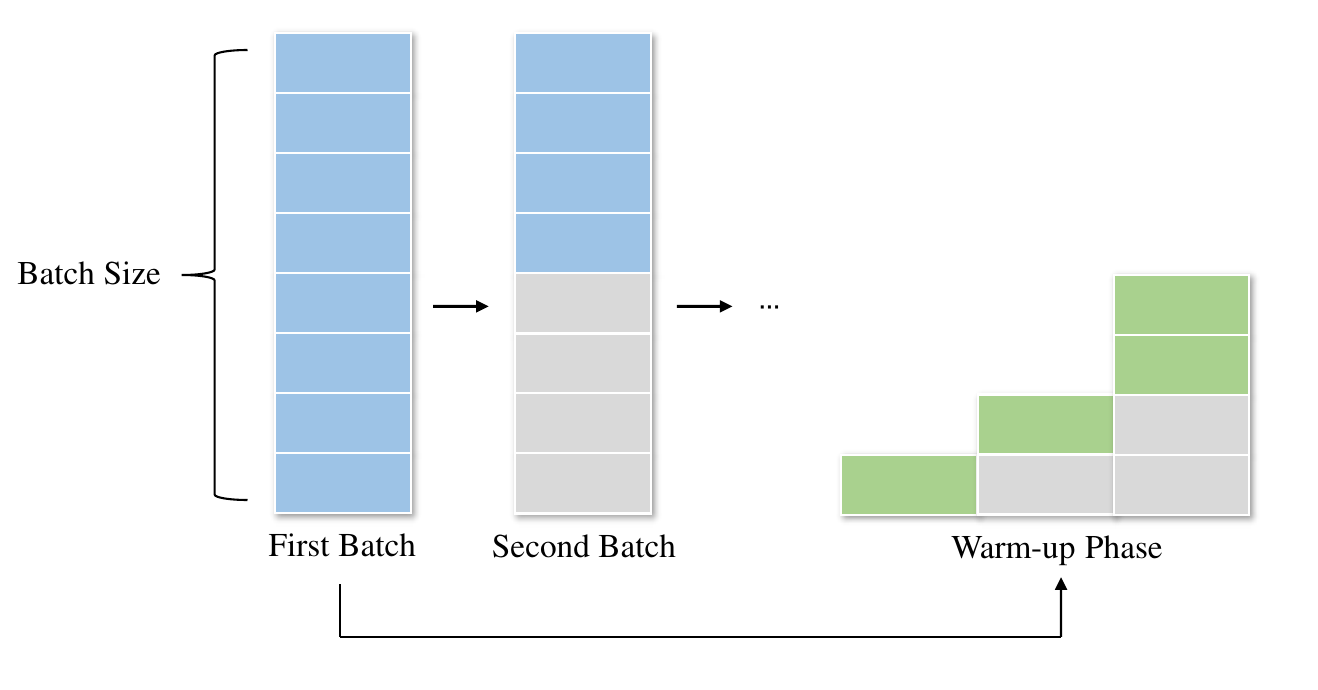}%
		\label{fig:method:warm up}}
	\caption{Figure (a) depicts the Root filter process, where lower-scoring triplets are directly filtered out. On the right, Figure (b) demonstrates Batch warm-up. In this process, the data from the initial batch undergoes decomposition, and its size gradually increases until it reaches a predetermined limit. And subsequent batches maintain a consistent size.}
	\label{fig:}
\end{figure*}

As depicted in Figure \ref{fig:method:root filter}, we possess the capability to exclude facts with scores inferior to that of the root element within the heap. This filtration procedure can be executed in parallel, resulting in a substantial acceleration in processing speed without overlooking any candidate facts.

\paragraph{warm-up} While the root filter significantly expedites the processing of subsequent batches, its effectiveness does not extend to the initial batch. We have observed that when dealing with a sizable batch size, the execution of the first batch experiences notable delays. To address this challenge and to maximize the utility of the root filter, we have introduced a warm-up schedule for the batch size within each mining process. Specifically, we initiate with a batch size of 1 and systematically increase it exponentially until it aligns with the intended batch size, as illustrated in Figure \ref{fig:method:warm up}. This approach is designed to mitigate the performance issues associated with larger batch sizes during the initial stages of the mining process.

\begin{figure}[!t]
  \renewcommand{\algorithmicrequire}{\textbf{Input:}}
  \renewcommand{\algorithmicensure}{\textbf{Output:}}
  \begin{algorithm}[H]
    \caption{Optimized Top-$k$ with SVF}\label{alg:optimized top-k}
    \begin{algorithmic}[1]
      \REQUIRE Known Facts $\mathcal{F}_{known}$, Visited Facts $\mathcal{F}_{visited}$, model $s(\cdot)$, mining maximum batch size $b_{m}^{max}$, class dict $\mathcal{D}$, number of candidate $n_c$
      \ENSURE Mined candidate knowledge $\mathcal{F}_{mined}$
      \STATE \textbf{Initialization:} Batch begin index $b_b\gets 0$, Batch size $b_{m} \gets  1$, min-heap $H$ with $n_c$ dummy elements $f_d$ that $s(f_d) \gets -\infty$
      \STATE $\psi \gets$ $GetSVF (\mathcal{F}_{known}, \mathcal{D})$ \quad // \textit{SVF}
      \STATE $Q\gets GetValidQueries (\mathcal{F}_{known},\psi)$
      \WHILE{$b_b \leq Q.size$}
        \STATE $q \gets Q[b_b: b_b + b_m]$
        \STATE $\mathcal{F}_c\gets GetCandidateFacts(q)$
        \STATE $\mathcal{F}'_{c} \gets \{f | f \in \mathcal{F}_{c}, s(f) > s(H.root)\}$ \quad // \textit{Root Filter}
        \FOR{triple $f\in \mathcal{F}'_{c}$}
            \IF{$f \notin \mathcal{F}_{visited}$ \rm{\textbf{and}} $s(f) > s(H.root)$}
                \STATE $H.replace(f)$
            \ENDIF
        \ENDFOR
        \STATE $b_m \gets max(b_m^{max}, 2*b_m)$ \quad // \textit{warm-up} 
        \STATE $b_b \gets b_b + b_m$
      \ENDWHILE
      \STATE $\mathcal{F}_{mined} \gets Set(H)$
    \end{algorithmic}
  \end{algorithm}
  \vspace*{-3mm}
\end{figure}

\subsubsection{Semantic Validity Filter}
\label{sec: svf}
In addition to delving into the mining process, we tackle the aspect of "what to mine" by introducing a pivotal component, the Semantic Validity Filter (SVF), which efficiently trims down the search space. Our inspiration for this derives from the recognition that not all combinations of entities and relations are valid~\cite{li2022star}. For instance, the head entity of the relation 'isCityOf' can only be the name of a 'city', and not a 'human'.

The SVF implementation entails acquiring class information for each entity and maintaining a set that documents each pairing of class and relation initially present in the initial known KG $\mathcal{F}_{known}$. Throughout the mining process, we exclusively entertain queries that align with the SVF criteria, effectively eliminating invalid queries and significantly reducing the search space. It's worth noting that in cases where certain entities lack class information during data collection, we abstain from imposing constraints on queries for these entities. We present the operational details of these modules within Algorithm \ref{alg:optimized top-k}.

\section{Experiment}

In this section, our discussion unfolds in a structured manner. We initiate by introducing the experimental setup, elaborated upon in Section \ref{sec:experiment setting}. Subsequently, we unveil the key findings in Section \ref{sec:main result}, culminating in a thorough analysis of the observed phenomena in Section \ref{sec:analysis}.

\subsection{Experiment Setting}
\label{sec:experiment setting}

\paragraph{Datasets} Our datasets are meticulously crafted, rooted in FB15k and WN18~\cite{rescal}. To maintain a consistent initial completion ratio, denoted as $\rho$, we partition the dataset into two distinct categories: the initial triples, $\mathcal{K}_{known}$, and the unexplored triples, $\mathcal{K}_{un}$. During this data partitioning process, we impose stringent constraints to ensure that undetermined entities and relations do not appear in $\mathcal{F}_{un}$. Comprehensive dataset statistics are thoughtfully presented in Table \ref{table:datasets}. It's noteworthy that we consciously opted against utilizing the FB15k-237~\cite{fb237} and WN18RR~\cite{conve} datasets, favoring a pursuit of more lifelike scenarios. Our primary research focus pivots not on the dataset's inherent complexity, but rather on the nuanced exploration of basic model performance within the context of real-world scenarios. 

\begin{table}[!t]
\centering
\resizebox{0.95\linewidth}{!}{
\begin{tabular}{c|ccccc}
        \hline
        Dataset & $|\mathcal{E}|$ & $|\mathcal{R}|$ & $|\mathcal{K}_{known}|$ & $|\mathcal{K}_{un} |$ & $\rho$\\
        \hline
        FB15k & 14,951 & 1,345 & 532,989 & 59,221 & 0.9\\
        WN18 & 40,943 & 18 & 106,009 & 45,433 & 0.7\\ 
        \hline
    \end{tabular}
    }
    \caption{Statistics of Two Benchmark Datasets.}
    \label{table:datasets}
\end{table}

\paragraph{Partition} In the process of partitioning the dataset, our primary objective is to ensure the comprehensive inclusion of all entities and relations within $\mathcal{K}_{known}$. This goal guides a meticulous procedure, during which we thoroughly scrutinize each triple, diligently recording the entities and relations that make an appearance. In the event that a triple is encountered, containing components that have not yet found a place in our records, it is promptly added to the scaffold set. The remaining triples undergo a division into two distinct segments, this division hinging on the predetermined ratio $\rho$. One segment becomes $\mathcal{K}_{un}$, while the other segment is integrated into the scaffold set to form $\mathcal{K}_{known}$. For an exhaustive and step-by-step elucidation of this partitioning process, we direct your attention to Algorithm \ref{alg:partition}.

\begin{figure}[htp]
  \renewcommand{\algorithmicrequire}{\textbf{Input:}}
  \renewcommand{\algorithmicensure}{\textbf{Output:}}
  \begin{algorithm}[H]
    \caption{Dataset Partition}\label{alg:partition}
    \begin{algorithmic}[1]
      \REQUIRE Total Facts $\mathcal{F}_{total}$, initial ratio $\rho$
      \ENSURE Known Facts $\mathcal{F}_{known}$, Unexplored Facts $\mathcal{F}_{un}$
      \STATE \textbf{Initialization:} Scaffold triples set $S\gets \phi$, visited entity set $S_e\gets  \phi$, visited relation set $S_r\gets \phi$
      \FOR{$(h,r,t)\in \mathcal{F}_{total}$}
        \IF{$h \notin S_e$ \rm{\textbf{or}} $t \notin S_e$ \rm{\textbf{or}} $r \notin S_r$}
            \STATE $S_e.add(h)$
            \STATE $S_e.add(t)$
            \STATE $S_r.add(r)$
            \STATE $S.add((h,r,t))$
        \ENDIF
      \ENDFOR
      \STATE $n\gets len(\mathcal{F}_{total})\ast\rho $
      \STATE $S_{remain}\gets \mathcal{F}_{total}-S$ 
      \STATE $\mathcal{F}_{un}\gets S_{remain}[:len(\mathcal{F}_{total})-n]$
      \STATE $\mathcal{F}_{known}\gets S_{remain}[len(\mathcal{F}_{total})-n:]+S$
      
    \end{algorithmic}
  \end{algorithm}
  \vspace*{-3mm}
\end{figure}

\paragraph{Baselines} Our study encompasses a comprehensive comparative analysis between our task and the well-established traditional structure-based models, which can be categorized into two primary groups: distance-based models and bilinear models. The distance-based models considered in this study are TransE~\cite{bordes2013translating}, RotatE~\cite{sun2018rotate}, and RotE~\cite{rote}. On the other hand, the bilinear models comprise RESCAL~\cite{rescal}, CP~\cite{cp}, ComplEx~\cite{complex}, QuatE~\cite{quate}, and UniBi~\cite{unibi}. UniBi stands out for its unique approach of normalizing both the modulus of the entity vector and the spectral radius of the relational matrix to 1.

\paragraph{Training}
The PKGC training process consists of three phases: (1) Hyperparameter optimization, (2) Model training, and (3) Model update. During the initial phase, we partition the set of known facts, $\mathcal{F}_{known}$, into training and validation subsets to perform hyperparameter optimization, following the approach established in prior SKGC research. In the second phase, the model is trained using the complete dataset of $\mathcal{F}_{known}$. In the final phase, the model is updated based on verified facts. This paper primarily focuses on the first two phases, with the third phase introduced in Section \ref{sec:continual learning}. In these first two phases, we adopt a reciprocal setting wherein we generate a new reciprocal triple $(t, r^{\prime}, h)$ for every $(h, r, t)$ in $\mathcal{F}_{known}$. To enhance the model's expressiveness, we introduce the DURA and Frobenius norm as regularization terms with a positive weight parameter, denoted as $\lambda > 0$. Descriptions about hyperparameters are in Appendix~\ref{app:hyperparameters}.

\begin{scriptsize}
\begin{equation}
    \mathcal{L}=-\sum\limits_{(h,r,t)\in \mathcal{F}_{train}}\log(\frac{\exp(s(h,r,t))}{\sum_{t^{\prime}\in \mathcal{E}}\exp(s(h,r,t^{\prime}))})+\lambda \cdot Reg(h,r,t)
    \label{con:loss}
\end{equation}
\end{scriptsize}

In Equation \ref{con:loss}, we include an additional scalar parameter $\gamma > 0$ after $s(\cdot)$ to inherit the setting from UniBi. We give details of the hyperparameter settings in Section~\ref{app:hyperparameters}.

\begin{figure}[!t]   
    \centering
    \vspace*{-3mm}
    \includegraphics[width=0.95\linewidth]{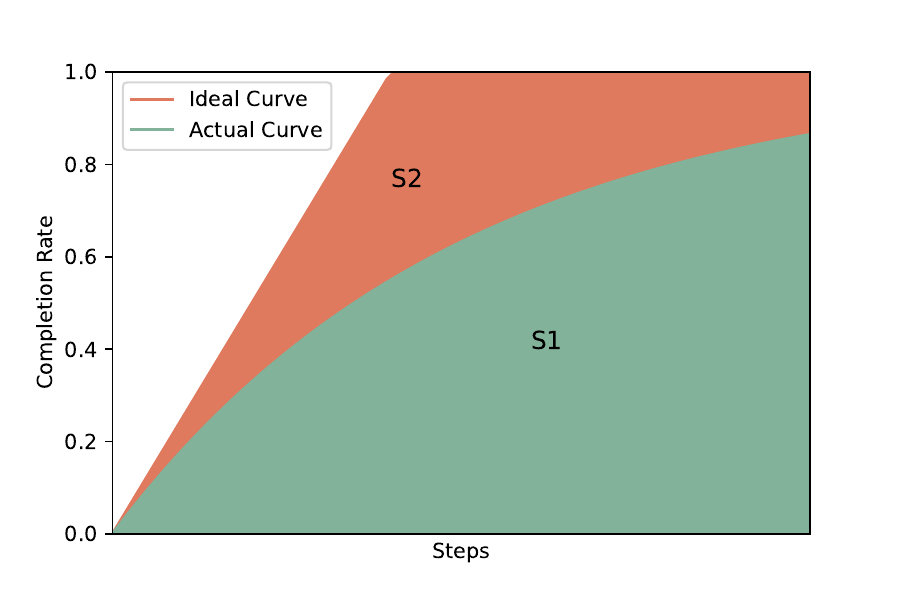}
    \caption{Illustration of how to calculate MOAR, which is the ratio of the area enclosed by the actual curve and the ideal curve, respectively. }
    \label{fig:metric moar}
\end{figure}

\begin{table*}[!t]
    \centering
    \resizebox{0.8\textwidth}{!}{
    \begin{tabular}{lcccccccc}
    \hline
    \multicolumn{1}{c}{} & \multicolumn{4}{c}{\textbf{WN18}}& \multicolumn{4}{c}{\textbf{FB15K}} \\
    Model & MRR & Hits@10  & MOAR & CR@50 & MRR & Hits@10 &  MOAR & CR@200  \\ \hline
TransE & 0.399 & 0.590 & 0.223 & 0.803 & 0.487 & 0.731 & 0.083 & 0.913 \\
CP & 0.563 & 0.687 & 0.515 & 0.920 & 0.529 & 0.749 & 0.588 & 0.987 \\
RotatE & 0.595 & 0.699 & \textbf{0.768} & \textbf{0.935} & 0.543 & 0.756 & 0.621 & 0.987 \\
RotE & 0.596 & 0.705 & \underline{0.765} & \textbf{0.935} & 0.540 & 0.761 & 0.317 & 0.945 \\
ComplEx & 0.561 & 0.692 & 0.630 & 0.927 & 0.547 & 0.765 & 0.623 & 0.986 \\
QuatE & 0.596 & 0.704 & 0.759 & 0.933 & 0.547 & 0.767 & 0.637 & 0.984 \\
RESCAL & 0.548 & 0.661 & 0.545 & 0.892 & 0.450 & 0.684 & 0.668 & 0.988 \\
UniBi-O(2) & \underline{0.598} & \underline{0.712} & 0.761 & \underline{0.934} & \textbf{0.550} & \textbf{0.771} & \underline{0.841} & \underline{0.994} \\
UniBi-O(3) & \textbf{0.599} & \textbf{0.716} & 0.762 & \underline{0.934} & \underline{0.548} & \underline{0.768} & \textbf{0.846} & \textbf{0.995} \\

\hline
    \end{tabular}}
    \caption{Results of different models on both previous and progressive metrics. Best results are \textbf{bold}, second ones are \underline{underlined}.}
    \label{tab:exp:main table}
\vspace{-5mm}
\end{table*}

During the mining and verification phases, a scatter chart is constructed to depict the relationship between the completion ratio and the number of steps, guided by $\mathcal{K}_{un}$. We designate the ultimate completion ratio achieved at step k as the critical metric CR@$k$. Furthermore, we employ the Area Under the Curve (AUC) of the ideal and practical completion curves to quantify an additional indicator, the Model-to-Oracle Area Ratio (MOAR), as illustrated in Figure~\ref{fig:metric moar} and computed using Equation \ref{con:moar calculation}. 

\begin{small}
\begin{equation}
    MOAR = \frac{S_1}{S_1+S_2}
    \label{con:moar calculation}
\end{equation}
\end{small}\newline

\vspace{-20pt}

We fix $k$ at $200$ for the FB15k dataset and at $50$ for the WN18 dataset, accounting for variations in the performance of the foundational models on these datasets.

\subsection{Main Result}
\label{sec:main result}
\begin{figure*}[htbp]
\centering
\subfloat[WN18.]{\includegraphics[width=3.0in]{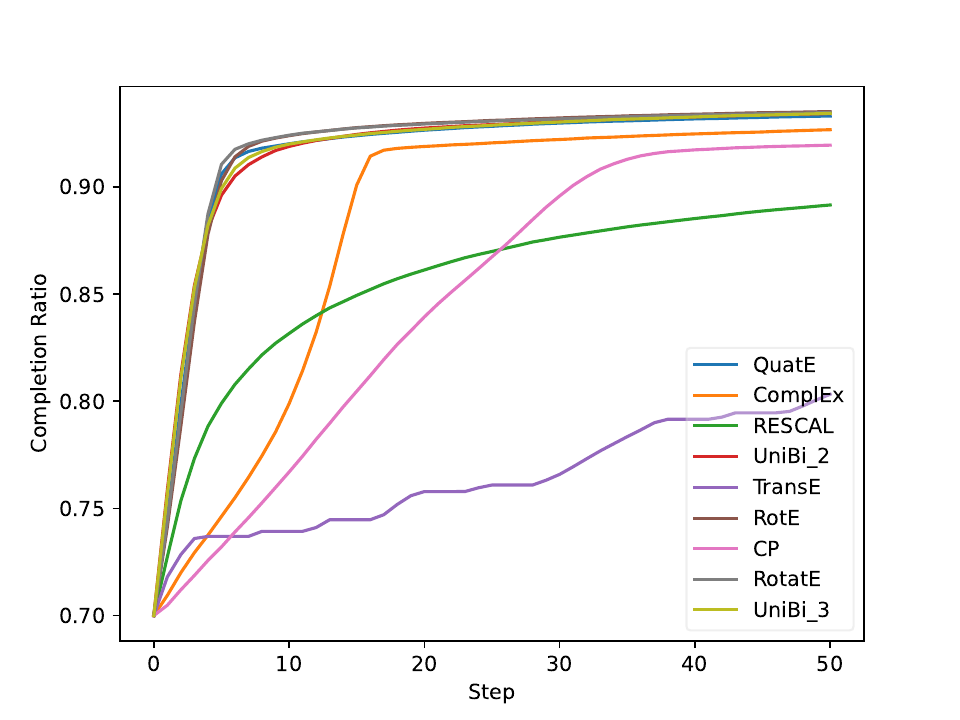}%
\label{fig:exp:dynamic wn}}
\hfil
\subfloat[FB15k.]{\includegraphics[width=3.0in]{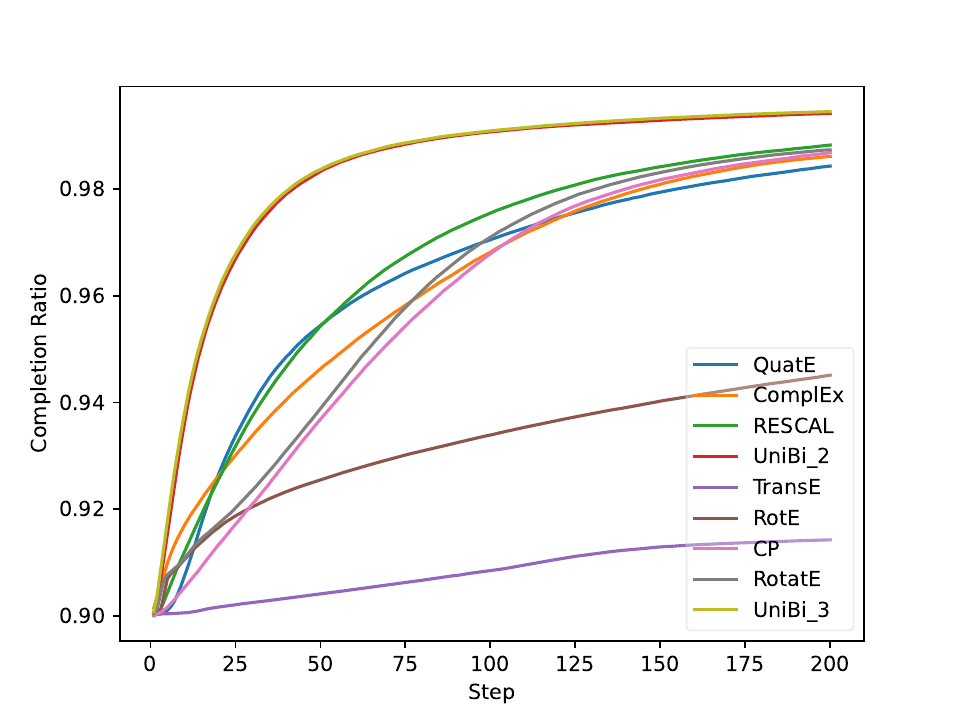}%
\label{fig:exp:dynamic fb}}
\caption{Figures illustrating the dynamic completion process for various models on the WN18 and FB15k datasets. The closer the trend of the curve is to the upper left, the more efficient the model is in performing the dynamic completion.  It is evident that UniBi-O(2) and UniBi-O(3) maintain a significant advantage on both datasets, while TransE performs poorly on both.}
\end{figure*}

The performance of the models on WN18 and FB15k is presented in Table \ref{tab:exp:main table}. Additionally, we provide a detailed illustration of the dynamics of the models on these datasets through Figure \ref{fig:exp:dynamic wn} and Figure \ref{fig:exp:dynamic fb} to enhance our understanding of the PKGC process.

Initially, we conduct separate analyses of the models' performance on SKGC and PKGC. Our findings indicate that most models exhibit strong performance on both datasets within the realm of SKGC, whereas this extends to only one dataset within PKGC. Notably, UniBi stands out as the sole model demonstrating commendable performance across all metrics and datasets.

Furthermore, we note that the performance of models in link prediction does not consistently align with their performance in PKGC. On one hand, the models' performance in link prediction and PKGC sometimes diverges. For instance, RESCAL exhibits the lowest Mean Reciprocal Rank (MRR) on FB15k but secures the third-best result in Completion Ratio (CR). On the other hand, models that perform similarly in link prediction may exhibit differences in PKGC performance. A case in point is the closeness in MRR between UniBi-O(3) and ComplEx (0.548 vs. 0.547), yet they demonstrate disparities in CR (0.995 vs. 0.986). Notably, QuatE and UniBi are the only models performing well across all metrics in both datasets.

\section{Analysis}
\label{sec:analysis}
In this section, we mainly discuss: 1) What are the key factors in PKGC, for which we have conducted a discussion related to the normalization~(\textbf{RQ1}); 2) The effectiveness of the acceleration module we proposed (\textbf{RQ2}); 3) Improving this task from the perspective of incremental learning, providing a feasible path for future research~(\textbf{RQ3}); 4) Whether PKGC can execute in low-resource situation~(\textbf{RQ4}).

\subsection{RQ1: How Normalization Work in PKGC?}

\begin{figure}[!t]
\centering
\vspace*{-4mm}
\subfloat[]{\includegraphics[width=0.5\linewidth]{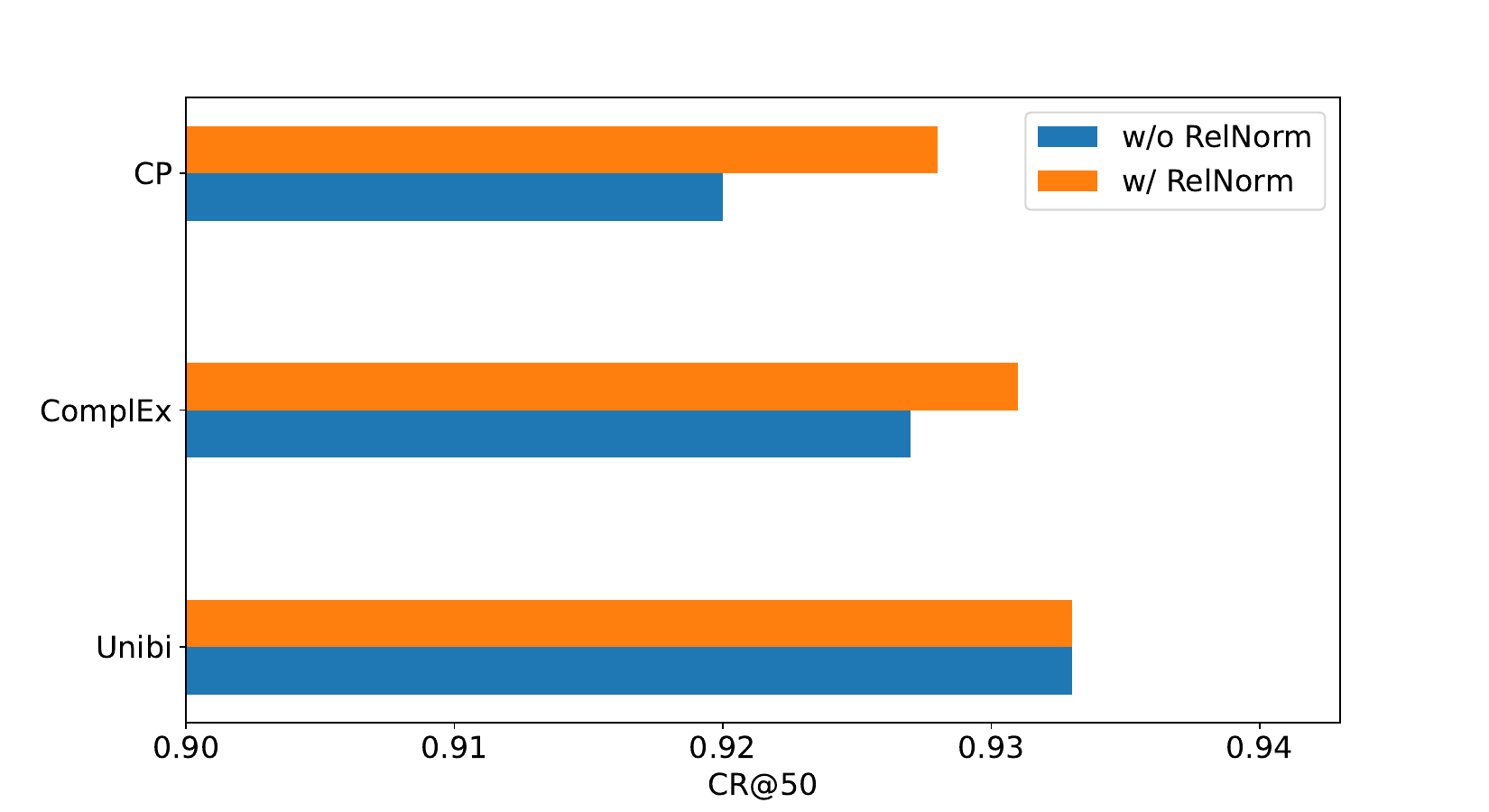}%
\label{fig:wn_norm}}
\subfloat[]{\includegraphics[width=0.5\linewidth]{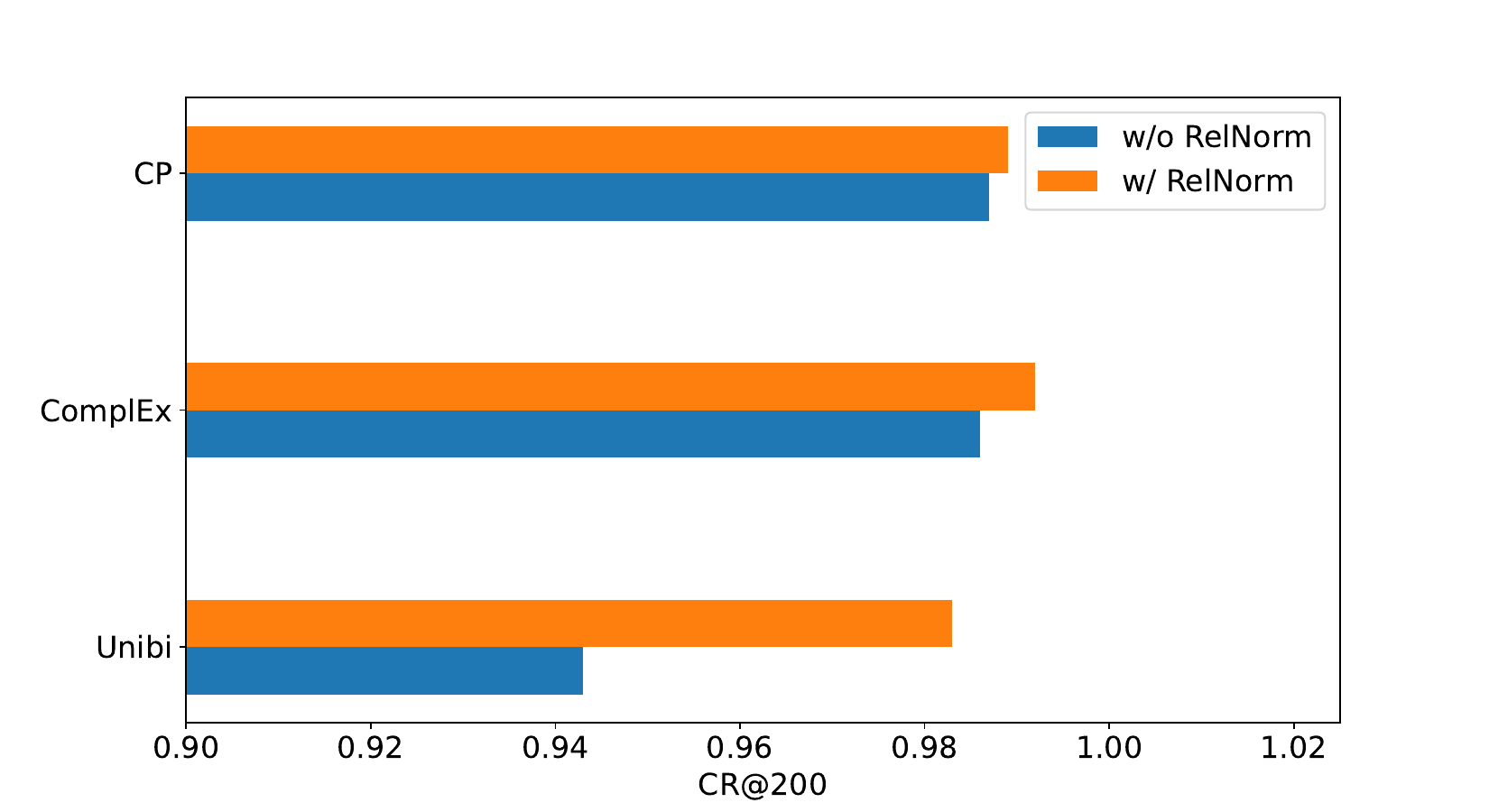}%
\label{fig:fb_norm}}
\caption{Ablation studies of relation normalization (RelNorm) on (a) WN18 and (b) FB15k. We utilize CP and ComplEx as examples.}
\label{fig:rel norm}
\vspace*{-3mm}
\end{figure}

Given UniBi's remarkable performance over other models, we delve into an investigation to uncover the factors contributing to its strength. UniBi's unique feature lies in its normalization of both entities and relations, which is unusual among bilinear-based models. We argue that UniBi's performance enhancement primarily stems from its ability to make triples comparable. To illustrate, consider two triples: $f_1 = (h, r_1, t)$ and $f_2 = (h, r_2, t)$, with corresponding scores $s(f_1) = a$ and $s(f_2) = b$, where $a > b$. Without normalization, determining the relative plausibility of $f_1$ and $f_2$ is challenging due to the unaccounted score range. For instance, consider the case where  $|\mathbf{e}| \leq 1$ and $s(\cdot) = \rho_r \cdot\mathbf{h}^\top \mathbf{t}$, confined within $[-\rho_r, \rho_r]$, with $\rho_{r_1} = 2a$ and $\rho_{r_2} = b$. Here, $s(f_2) = b$ achieves the highest score under relation $r_2$, while $s(f_1) < 2a$ still has potential for improvement. Thus, $f_2$ appears more plausible than $f_1$. UniBi distinguishes itself from other bilinear-based models by ensuring that $s(\cdot)$ falls within $[-1, 1]$ for any triple, enabling more rational comparisons between relations. We present CP and ComplEx as case studies, showing that both models improve their performance with relation normalization, as shown in Figure \ref{fig:rel norm}. We suggest that this advantage becomes more apparent when the number of triples per relation is limited, shedding light on why UniBi outperforms on FB15k but not on WN18.

\subsection{RQ2: Ablation on Acceleration Modules w.r.t. Time cost}

\begin{table}[!t]
    \centering
    \resizebox{0.95\linewidth}{!}{
    \begin{tabular}{cccrr}
    \hline
    \multicolumn{3}{c}{Module}& \multicolumn{2}{c}{Efficiency} \\
    Root Filter & warm-up & SVF & Time & Speed-up\\ 
    \hline
     & && $\approx 243d14h$ & 1$\times$\\
     & &  \checkmark & $\approx 32d4h$& 7.6$\times$\\
     \checkmark & & & $2h10m52s$ & 2,698.7 $\times$\\
     \checkmark & & \checkmark& $2h7m33s$ & 2,763.7$\times$\\
     \checkmark & \checkmark & & $3m38s$ & 92,337.2$\times$ \\
     \checkmark & \checkmark & \checkmark & $50s$ & 421,057.6$\times$ \\
\hline
    \end{tabular}
    }
    \caption{Ablation studies on proposed acceleration modules. The results of first two rows are estimated.}
    \label{tab:exp:ablation on proposed acceleration modules}
\end{table}

We commence by demonstrating the efficacy of two acceleration modules proposed for knowledge mining. Results in Table \ref{tab:exp:ablation on proposed acceleration modules} reveal that all three modules contribute significantly to expediting the mining process, with the root filter exhibiting the most substantial speedup, exceeding 2,000 times. More discussion are in Appendix~\ref{app:ablation}.

\subsection{RQ3: Incremental Aspect on PKGC}
\label{sec:continual learning}

In order to harness freshly validated knowledge, we employ the approach of incremental learning through retraining and fine-tuning, incorporating both existing and recently confirmed facts into the ongoing training process. We conduct incremental model training at a specified frequency denoted as $\Delta s$~($\Delta s = 5$ in our setting). During retraining, our training data encompasses $\mathcal{F}_{known}$, which comprises the verified facts from the most recent $\Delta s$ steps denoted as $\mathcal{F}_{new}$. Conversely, during fine-tuning, our training is exclusively focused on the facts in $\mathcal{F}_{new}$ from the recent $\Delta s$ steps. Additionally, we introduce an additional term to ensure that the previously acquired entity representations undergo moderate alterations~(Formulated expression in Appendix~\ref{app:incremental_training}).

\begin{table}[!t]
    \centering
    \resizebox{\linewidth}{!}{
    \begin{tabular}{lcccccc}
    \hline
    \multicolumn{1}{c}{} & \multicolumn{2}{c}{\textbf{Origin}}& \multicolumn{2}{c}{\textbf{Retraining}} & \multicolumn{2}{c}{\textbf{Fine-tuning}} \\
    Model  & MOAR & CR@50 & MOAR & CR@50 & MOAR & CR@50 \\ \hline
CP & 0.515 & 0.920 & 0.570 & 0.921 & 0.570 & 0.920 \\
RotatE & \textbf{0.768} & \textbf{0.935} & \textbf{0.766} & \underline{0.935} & 0.753 & 0.928 \\ 
ComplEx & 0.630 & 0.927 & 0.631 & 0.927 & 0.637 & 0.928 \\
QuatE & 0.759 & 0.933 & \underline{0.765} & \underline{0.935} & \textbf{0.764} & \textbf{0.933} \\
UniBi-O(2) & \underline{0.761} & \underline{0.934} & \textbf{0.766} & \textbf{0.936} & \underline{0.758} & \underline{0.931} \\ 

\hline
    \end{tabular}}
    \caption{Results of different strategies on WN18 dataset.}
    \label{tab:exp:continual learning results}
\vspace{-2mm}
\end{table}

The outcomes, as depicted in Table \ref{tab:exp:continual learning results}, pertaining to the WN18 dataset, reveal that retraining often proves effective in bolstering performance, whereas fine-tuning may have adverse consequences. During progressive mining, at every stage, the recently acquired facts consistently receive high rankings. This implies that the model inherently possesses confidence in accurately identifying these factual triplets. Consequently, the impact of incorporating these new data into incremental learning is relatively modest. As emphasized in the context of active learning~\cite{active2}, it is crucial to emphasize samples near the decision boundary, particularly those involving low-ranking factual triplets.

\subsection{RQ4: Low-resource PKGC}
In this section, we explore low-resource PKGC to showcase its extensive practical applications. To achieve this, we reduce $\rho$, resulting in training the model on a smaller $\mathcal{K}_{known}$ while exploring the knowledge within a larger $\mathcal{F}_{un}$.Table \ref{tab:app:low resource cr} demonstrates that even in a low-resource scenario, UniBi ultimately attains reasonably satisfactory completion rates. On the WN18 dataset, when $\mathcal{K}_{known}$ holds only 30\% of the knowledge, UniBi achieves nearly 60\% completeness after 50 rounds of mining. With an initial knowledge base comprising 50\%, this figure increases to almost 80\%. On the FB15k dataset, after 200 completion rounds, UniBi enhances a knowledge base with a 50\% completeness rate to nearly 90\%. Similarly, concerning the CR@$k$ metric, UniBi consistently retains its ranking position in comparison to the CP model.The results demonstrate that even in resource-constrained settings, PKGC consistently yields favorable outcomes, establishing it as a pragmatic experimental framework.

\begin{table}[!t]
    \centering
    \resizebox{\linewidth}{!}{
    \begin{tabular}{lcccccc}
        \hline
        \multicolumn{1}{l}{} & \multicolumn{3}{c}{\textbf{WN18}}& \multicolumn{3}{c}{\textbf{FB15K}}\\
         Model & $\rho=0.3$ & $\rho=0.5$ & $\rho=0.7$ & $\rho=0.5$ & $\rho=0.7$ & $\rho=0.9$\\
         \hline
         CP & 0.510 & 0.758 & 0.920 & 0.841 & 0.934 & 0.987\\
         RotatE & 0.560 & \textbf{0.806} & \textbf{0.935} & 0.850 & 0.946 & 0.987 \\
         RESCAL & 0.521 & 0.795 & 0.892 & 0.821 & 0.932 & 0.988\\
         UniBi-O(2) & \textbf{0.579} & 0.797 & \underline{0.934} & \underline{0.894} & \underline{0.963} & \underline{0.994}\\
         UniBi-O(3) & \underline{0.578} & \underline{0.799} & \underline{0.934} & \textbf{0.907} & \textbf{0.964} & \textbf{0.995} \\
         \hline
    \end{tabular}}
    \caption{CR@$k$ of UniBi-O(2) and CP on low-resource WN18 and FB15k. ($k=50$ for WN18 and $k=200$ for FB15k, MOAR results are in Appendix~\ref{app:sec:supp_low_resource}).}
    \label{tab:app:low resource cr}
\end{table}

\section{Conclusion}

In this paper, we introduce a novel task called PKGC, which offers increased realism and a fundamental distinction by incorporating a finite verifier, and shifts the status of the model to that of a candidate knowledge proposer.In order to make the task more cost-effective, we have devised an optimized top-$k$ algorithm that integrates with the semantic validity filter to significantly expedite the mining process in PKGC. Furthermore, We conduct a comprehensive series of experiments to demonstrate the performance of baseline models and to analyze factors that distinguish them. Our findings indicate that previous metrics and knowledge are insufficient for accurately predicting and explaining phenomena within PKGC. Additionally, We have preliminarily explored the effectiveness of some incremental learning methods within the context of PKGC, highlighting the potential for further development of the task.

Hence, we conclude that PKGC not only serves as a novel benchmark but also represents a valuable avenue for investigating models' behavior in a more direct reflection of real-world scenarios. Future research directions include the pursuit of a more effective model and the exploration of methods to adaptively integrate incremental learning into the PKGC framework.

\bibliography{anthology,custom}
\bibliographystyle{acl_natbib}

\newpage
\appendix
\label{sec:app}
\section{Comparison between PKGC and SKGC}
\label{sec:comparison to surrogate completion}
\subsection{Connectivity Perspecitve}
In this section, we delve into an examination of the task disparities with a specific focus on their connectivity attributes. This analysis involves treating each fact as a vertex and each comparison between facts as an edge, thereby evaluating the connectivity of the resultant graph. As elucidated in Figure \ref{fig:app:skcg pkcg connected graph}, our findings reveal the following distinctions: (a) triple classification fails to establish any form of graph due to its exclusive reliance on comparisons between facts and a predefined threshold, (b) link prediction yields outcomes for distinct connected components associated with different queries, and (c) mining leads to the creation of a connected graph.

It is essential to note that in cases where a Knowledge Graph regresses into a homogenous graph characterized by a single relation, there is inevitably only one connected component, which ensures the graph's connectedness. This observation reinforces our belief that, link prediction emerges as a more suitable task in the context of Graph Neural Networks (GNN) instead of KGE.

\subsection{Comparison Perspective}

Within this section, our objective is to elucidate the disparities inherent in the evaluation tasks of SKGC and PKGC. We focus on the scope of comparison, in addition to recognizing the apparent distinction in progressiveness.

\begin{figure*}[htbp]
	\centering
    \vspace{-10pt}
	\subfloat[Scope of connectivity.]{\includegraphics[width=4.0in]{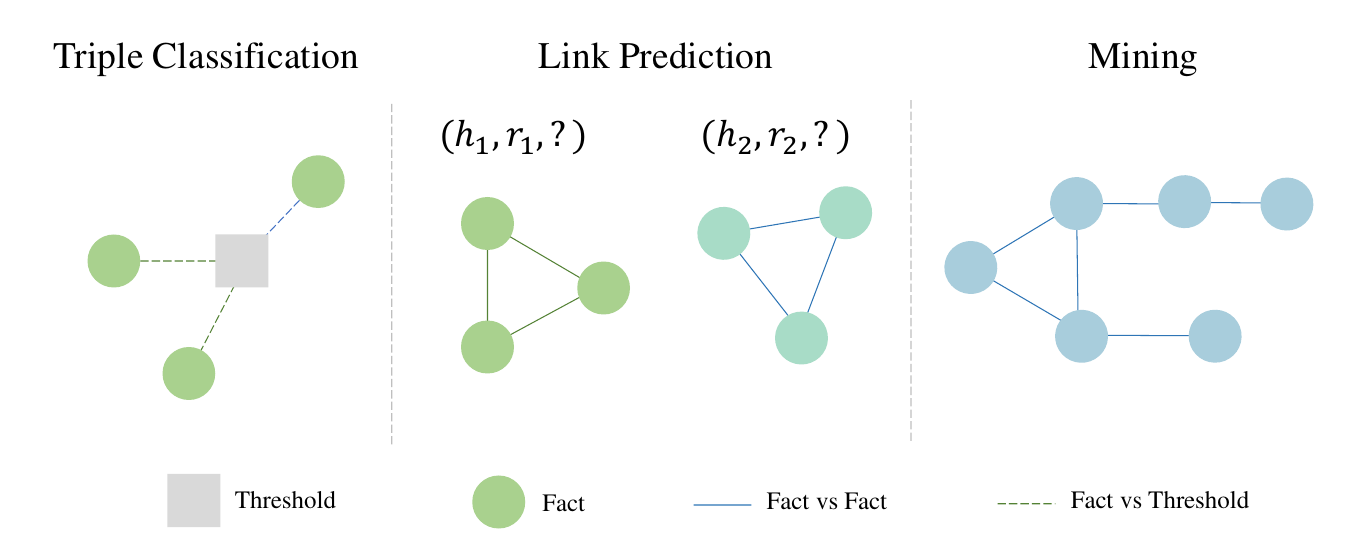}%
		\label{fig:app:skcg pkcg connected graph}}
	\hfil
	\subfloat[Scope of comparison]{\includegraphics[width=2.2in]{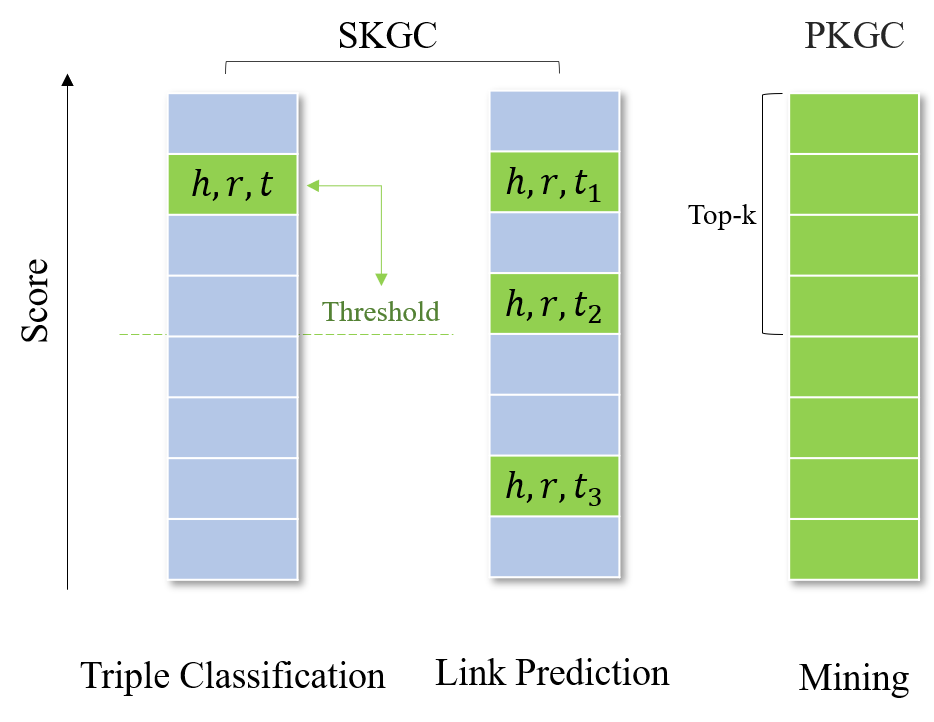}%
		\label{fig:skgc vs pkgc in task}}
	\caption{Two perspectives are used to explain the differences between PKGC and SKGC. Figure (a) provides clarification from the perspective of connectivity. Figure (b) presents it based on the scope of comparison. We use columns to denote all possible facts, and mark facts to be compared with green. }
	\label{fig:comparison_between_skgc_pkgc}
\end{figure*}

SKGC encompasses two fundamental tasks: triple classification and link prediction. As depicted in Figure \ref{fig:skgc vs pkgc in task}, triple classification entails the comparison of each fact with a predefined threshold, while link prediction involves juxtaposing facts within the same query $(h,r)$. However, what distinguishes these tasks is the absence of a mechanism for comparing facts between different queries. Consequently, in these tasks, facts remain isolated and segmented in terms of comparison.

Conversely, the mining process within PKGC mandates a global scope of comparison, permitting the evaluation of any two facts during the process of identifying the top-k facts. This expansive scope is not an inherent feature of SKGC tasks but rather arises as a necessity due to the involvement of realistic verifiers. Given the constraints imposed by limited verifiers\footnote{In cases where verifiers are limitless, they could exhaustively enumerate all possible facts, rendering the need for any model obsolete.}, a mining process becomes a vital component, serving to deliver the most valuable candidates.

In light of these considerations, we contend that the scope represents another pivotal distinction between SKGC and PKGC, complementing the aspect of progressiveness. It is crucial to recognize that this distinction cannot be overcome merely by expanding the number of test samples in SKGC tasks. Even in scenarios where these tasks undergo exhaustive testing, encompassing all possible facts, their comparisons would continue to be confined to individual or partial assessments, failing to adopt a holistic perspective.

\section{Supplementary Details in Incremental Training}
\label{app:incremental_training}

It is noteworthy that we do not apply a uniform regularization term for relation representations, as different models handle relations in diverse ways. 
\begin{equation}
    Reg_c(h,r,t) = ||\mathbf{h}_{new} - \mathbf{h}_{old}|| + ||\mathbf{t}_{new} - \mathbf{t}_{old}||,
\end{equation}

In this context, $\mathbf{e}_{new}$ denotes the updated entity embedding, whereas $\mathbf{e}_{old}$ signifies the entity embedding prior to the update.We have detailed the exact loss equations for retraining and fine-tuning in Equations \ref{equ:continue:retrain} and \ref{equ:continue:finetune}.

\begin{small}
\begin{equation}
\begin{aligned}
    \mathcal{L}_{retrain}&=-\sum\limits_{(h,r,t)\in \mathcal{F}_{known}}\log(\frac{\exp(s(h,r,t))}{\sum_{t^{\prime}\in \mathcal{E}}\exp(s(h,r,t^{\prime}))}) \\
     &+ \lambda \cdot Reg(h,r,t) + \mu \cdot Reg_c(h, r, t) 
    \label{equ:continue:retrain}
\end{aligned}
\end{equation}
\end{small}

\begin{small}
\begin{equation}
\begin{aligned}
    \mathcal{L}_{new}&=-\sum\limits_{(h,r,t)\in \mathcal{F}_{new}}\log(\frac{\exp(s(h,r,t))}{\sum_{t^{\prime}\in \mathcal{E}}\exp(s(h,r,t^{\prime}))}) \\
    &+\lambda \cdot Reg(h,r,t) + \mu \cdot Reg_c(h, r, t)
    \label{equ:continue:finetune}
\end{aligned}
\end{equation}
\end{small}

\section{Hyperparameters in Training}\label{app:hyperparameters}

To fully demonstrate the model's potential, we systematically explore hyperparameters, as listed in Table \ref{tab:exp:search hyperparameter}, utilizing the verification data. Of particular significance are the hyperparameters related to regularization, where weights are chosen from a range of values, specifically {0.0, 0.001, 0.003, 0.005, 0.01}. The ultimate selection of certain hyperparameters is documented in Table \ref{tab:exp:hyperparameter search table}. Table \ref{tab:exp:regularization weight table} displays the MOAR results for the RotatE and QuatE models, achieved through the utilization of DURA and F2 regularization techniques.

\begin{table*}[htp]
    \centering
    \begin{tabular}{lcccccccc}
    \hline
    \multicolumn{1}{c}{} & \multicolumn{4}{c}{\textbf{WN18}}& \multicolumn{4}{c}{\textbf{FB15K}} \\ 
    Model & 0.001 & 0.003 & 0.005 & 0.01 & 0.001 & 0.003 &  0.005 & 0.01 \\ \hline
    RotatE + F2 & 0.763 & 0.764 & 0.768 & 0.766 & 0.597 & 0.599 & 0.609 & 0.621 \\ 
    RotatE + DURA & 0.759 & 0.760 & 0.760 & 0.763 & 0.616 & 0.614 & 0.607 & 0.571 \\
    QuatE + F2 & 0.757 & 0.759 & 0.757 & 0.757 & 0.637 & 0.628 & 0.605 & 0.583 \\
    QuatE + DURA & 0.758 & 0.757 & 0.752 & 0.756 & 0.629 & 0.582 & 0.560 & 0.509\\
 
\hline
    \end{tabular}
    \caption{Results of MOAR obtained from RotatE and QuatE in different regularization weights.}
    \label{tab:exp:regularization weight table}
\end{table*} 

\begin{table}[htp]
    \centering
    \caption{Details of hyperparameters.}
    \resizebox{\linewidth}{!}{
    \begin{tabular}{lc}
    \hline
    Hyperparameters & Values \\ \hline
    Batch size(Pretrain) & 1000 \\
    Learning rate(Pretrain) & 0.001 \\
    Learning rate(Retrain \& Finetune) & 0.001 \\
    Embedding dimension & 500 \\
    Regularization weight & \{0.0, 0.001, 0.003, 0.005, 0.01\} \\
    Update frequency(Retrain \& Finetune) $\Delta s$ & 5 \\
    Max epoch(Pretrain) & 100 \\
    Epoch(Retrain \& Finetune) & 20 \\
    Max batch size of Root Filter $b_m^{max}$ & 10000 \\
    Regularization Term  & \{F2, DURA\} \\

\hline
    \end{tabular}}
    \label{tab:exp:search hyperparameter}
\end{table} 

\begin{table}[htp]
    \centering
    \caption{Optimal hyperparameter setting after search.}
    \resizebox{\linewidth}{!}{
    \begin{tabular}{lcccc}
    \hline
    \multicolumn{1}{c}{} & \multicolumn{2}{c}{\textbf{WN18}}& \multicolumn{2}{c}{\textbf{FB15K}} \\
    Model & Regularization & Weight & Regularization & Weight \\ \hline
    TransE & F2 & 0.003 & F2 & 0.003 \\ 
    CP & F2 & 0.001 & F2 & 0.0 \\
    RotatE & F2 & 0.005 & F2 & 0.01 \\
    RotE & F2 & 0.01 & F2 & 0.01\\
    ComplEx & DURA & 0.001 & DURA & 0.001\\
    QuatE & F2 & 0.003 & F2 & 0.001\\
    RESCAL & F2 & 0.001 & F2 & 0.003\\
    UniBi-O(2) & DURA & 0.01 & DURA & 0.01\\
    UniBi-O(3) & DURA & 0.01 & DURA & 0.01\\
 
\hline
    \end{tabular}}
    \label{tab:exp:hyperparameter search table}
\end{table} 

\section{Supplementary Details of Proposed Modules Ablation Study}
\label{app:ablation}

\subsection{Ablation on Hyperparameters in Mining}
\begin{figure}[!t]
    \centering
    \hspace*{-3mm}
    \subfloat[]{
    \includegraphics[width=1.6in]{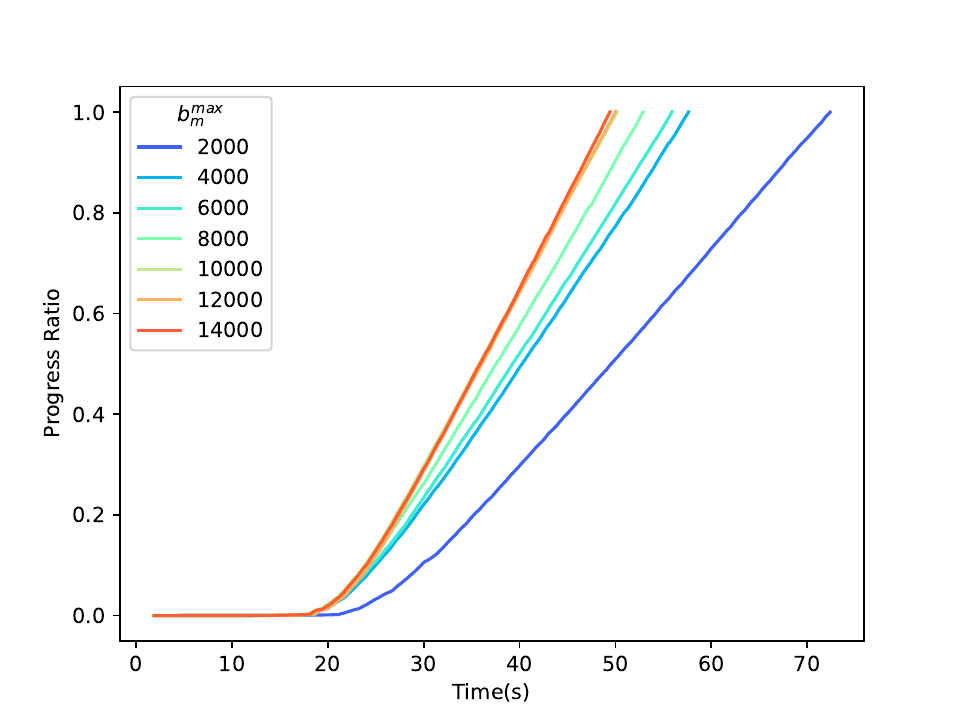}
    \label{fig:exp:Mining progresses for different batch size}
    }\hspace*{-3mm}
    \subfloat[]{
    \includegraphics[width=1.6in]{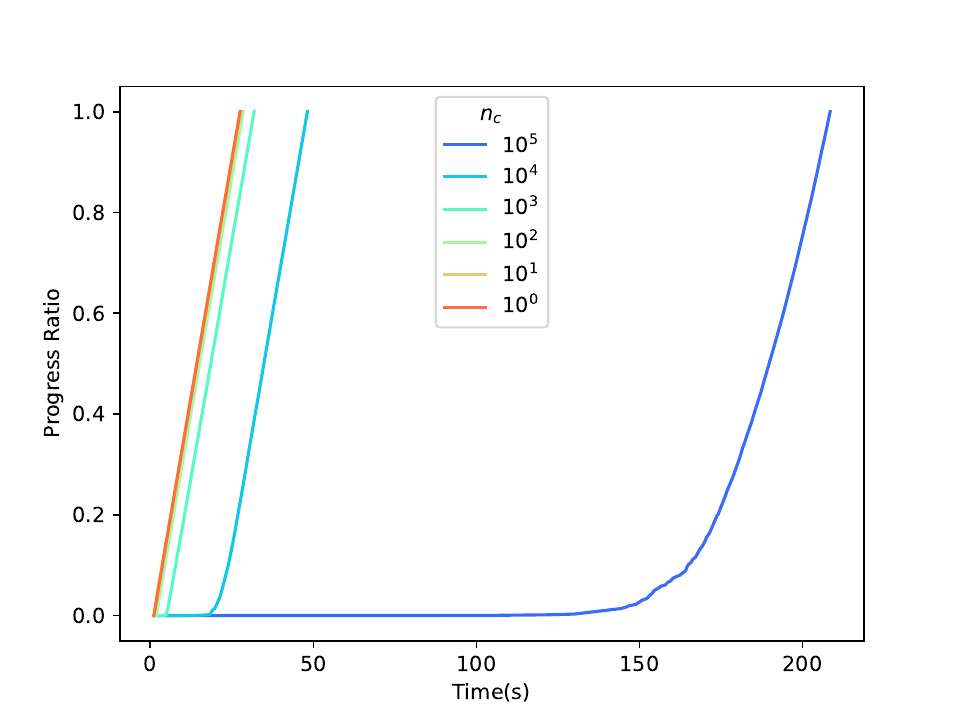}
    \label{fig:exp:mining time verification nums}
    }
    \caption{Ablation studies on (a) maximum mining batch size $b_m^{max}$ and (b) number of candidate $n_c$ in mining process. }
\end{figure}

In this section, we examine the influence of hyperparameters on the mining process, with a particular focus on the maximum batch size for mining, denoted as $b_m^{max}$, and the number of candidates, represented as $n_c$. Our results, as depicted in Figure \ref{fig:exp:Mining progresses for different batch size}, demonstrate that an increase in $b_m^{max}$ leads to accelerated completion. Nevertheless, we also note that this acceleration effect saturates as the batch size reaches a specific threshold. We attribute this saturation to the interplay between CPU computations and communication with the GPU.

\begin{figure}[!t]
    \centering
    \hspace{-7mm}
    \subfloat[]{
    \includegraphics[width=1.5in]{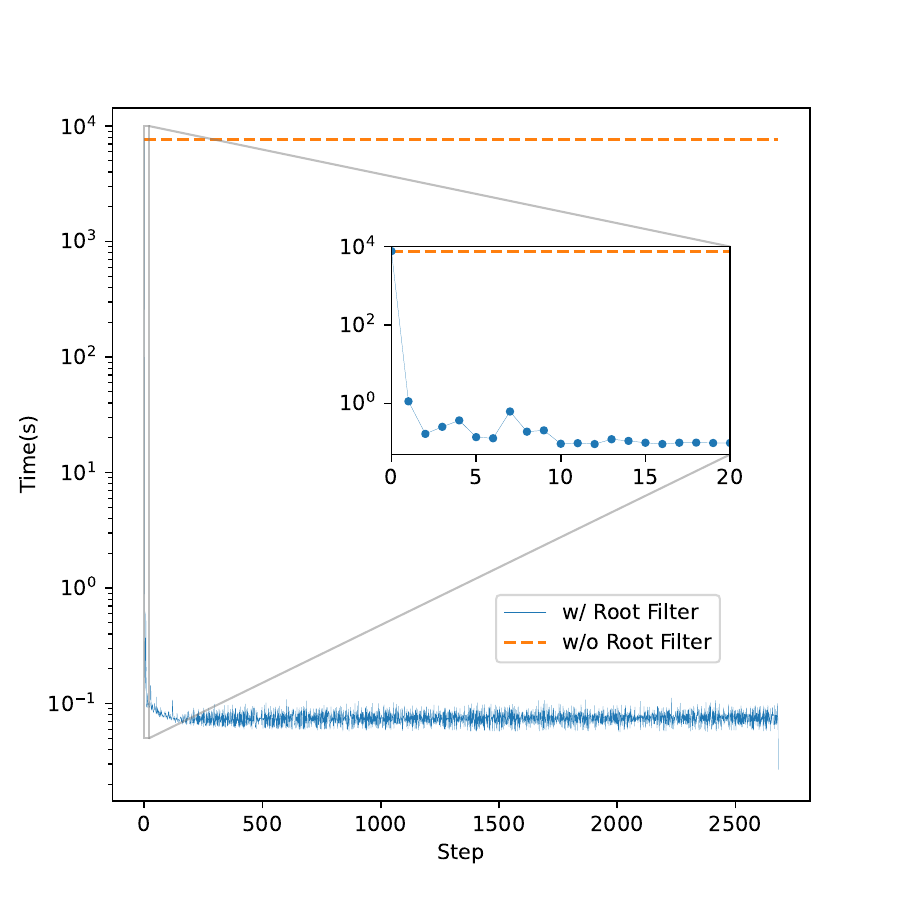}
    \label{fig:exp:The effectiveness of Root Filter}
    }
    \subfloat[]{
    \includegraphics[width=1.5in]{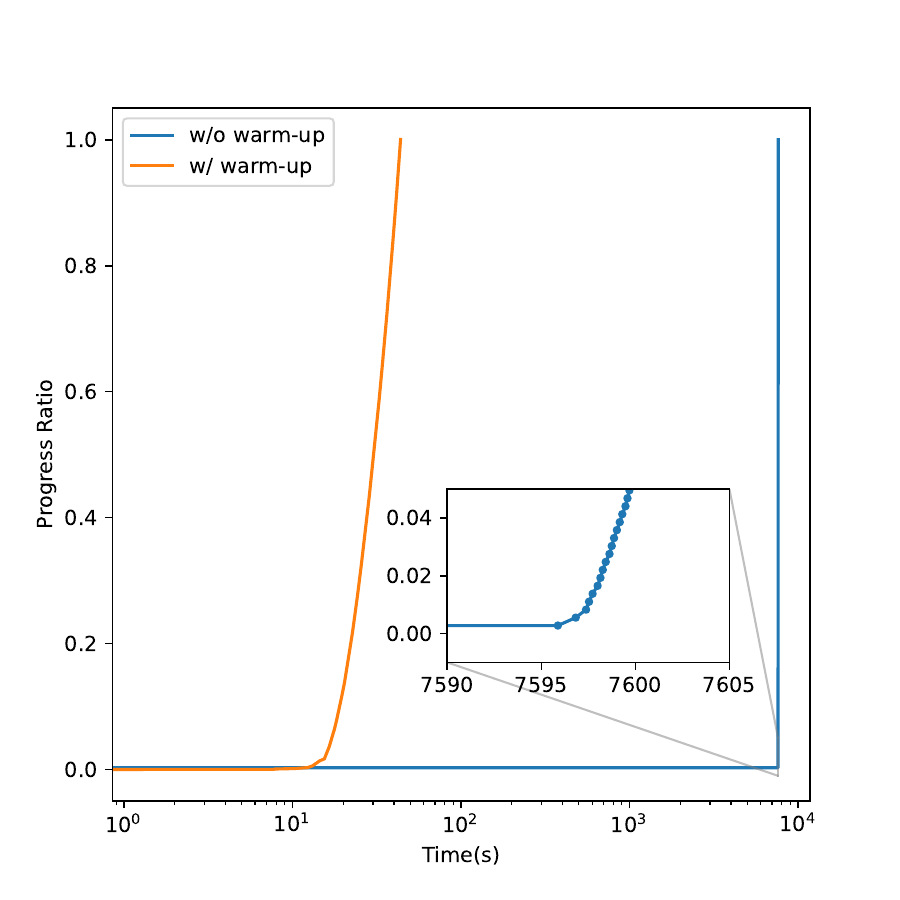}
    \label{fig:exp:effective warm up}
    }
    \caption{Ablation studies of (a) root filter and (b) warm-up modules. The dot lines is estimated.}
\end{figure}

\subsection{Detailed Ablation Studies w.r.t Time Cost}

we observe that the root filter alone does not suffice to optimize the process. Furthermore, we have discovered that the SVF module fails to yield significant improvements in the absence of the warm-up module. This limitation is attributed to the fact that the naive root filter consumes the majority of time in the initial batch, as depicted in Figure \ref{fig:exp:The effectiveness of Root Filter}. To address this, Figure \ref{fig:exp:effective warm up} illustrates how the warm-up module effectively mitigates congestion in the first batch, thus highlighting its importance.

\subsection{Efficiency of Optimized Top-$k$}
Due to the accumulation of visited facts suggested by Algorithm \ref{alg:pkgc} line 6, the speed of the algorithm decreases with the number of steps. Specifically, as the number of steps increases, the amount of known knowledge increases correspondingly, while the heap decreases its update frequency. Consequently, the efficiency of our algorithm decreases with the number of steps.

\begin{figure}[!t]
    \centering
    \subfloat[]{
    \includegraphics[width=1.55in]{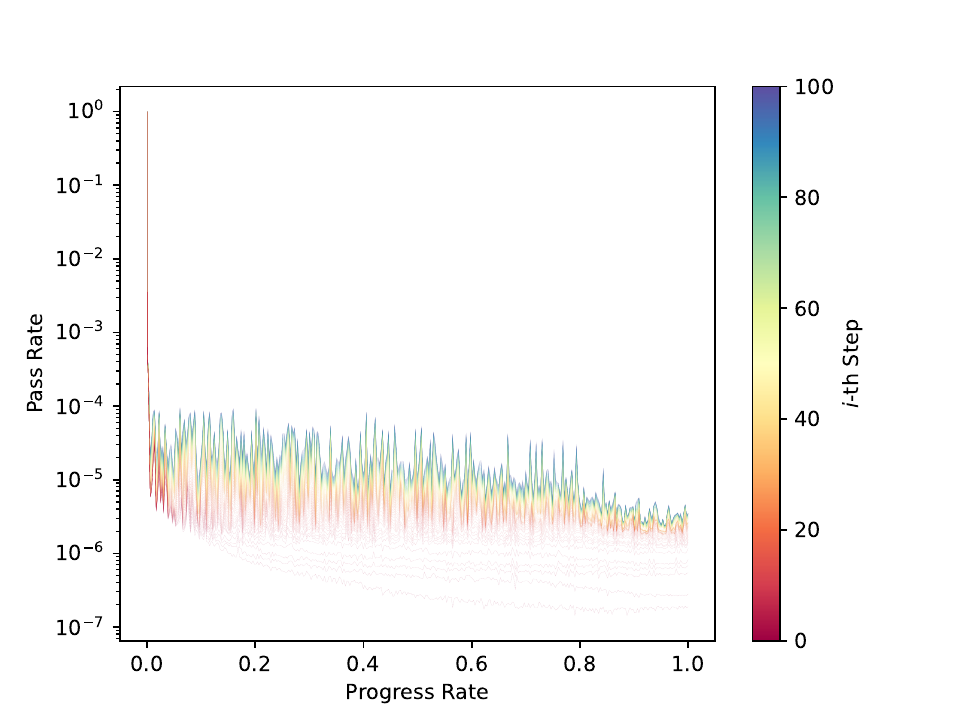}
    \label{fig:exp:Pass rate in different progresses and steps}
    }
    \hspace{-7mm}
    \subfloat[]{
    \includegraphics[width=1.55in]{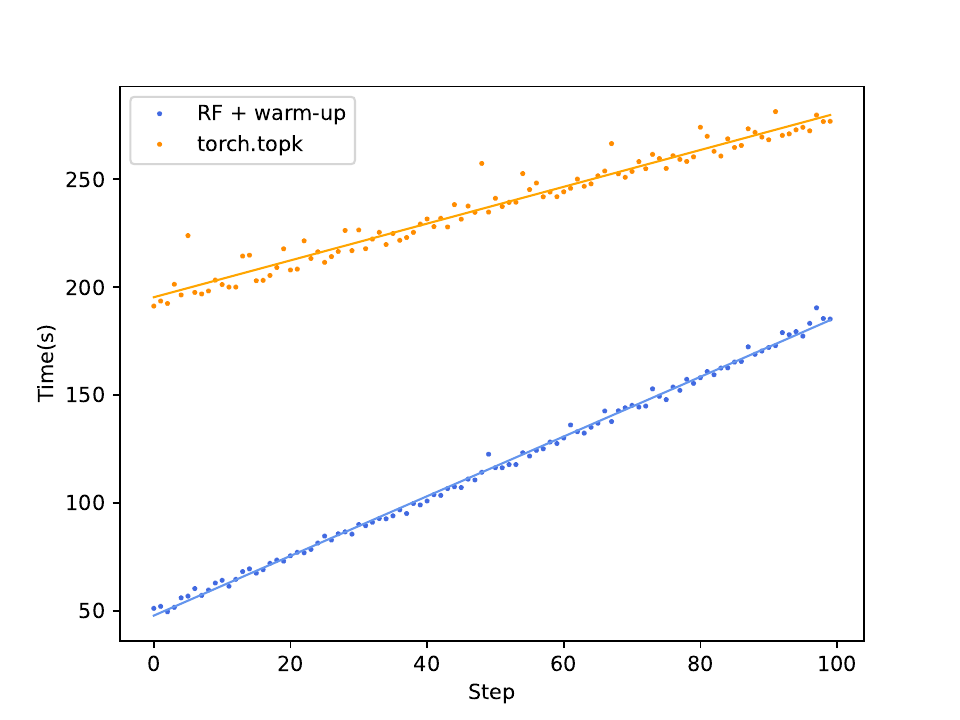}
    \label{fig:exp:mining time}
    }
    \caption{Ablation studies on pass rate and time consumption in different step. (a) Pass rate in different progresses and steps. The color represents the step, more small the step, more \textcolor{red}{red} the line, and more large the step, more \textcolor{blue}{blue} the line.  (b) Mining time during the steps. Here we also demonstrate that the combination of root filter (RF) and warm-up is faster than the implementation of topk on torch.}
\end{figure}

As depicted in Figure \ref{fig:exp:Pass rate in different progresses and steps}, the pass rate, representing the percentage of triples retained and sorted, exhibits two distinct trends. On one hand, the pass rate declines with the progression of mining in each step, influenced by the growing heap root that filters more triples. Conversely, the pass rate increases with the number of steps, as a substantial number of triples have been visited in prior steps, resulting in less frequent heap updates. Notably, the algorithm's effectiveness persists despite the declining pass rate.

Despite a decrease in the quantity of filtered triples with increasing steps, the algorithm's execution time demonstrates a linear growth, maintaining an acceptable range of performance, as depicted in Figure \ref{fig:exp:mining time}. Furthermore, our algorithm exhibits superior performance compared to the naive and incomplete implementation reliant on $torch.topk$.

In summary, our experiments provide compelling evidence for the algorithm's effectiveness. It is crucial to acknowledge that the observed variations with the number of steps are contingent upon the assumption that the number of candidates $n_c$ is considerably larger than the size of the knowledge graphs. In the context of a larger knowledge graph with the same $n_c$, these variations would be less pronounced.

\subsection{Performance after SVF}
\begin{figure*}[!t]
\centering
\subfloat[WN18]{\includegraphics[width=3.0in]{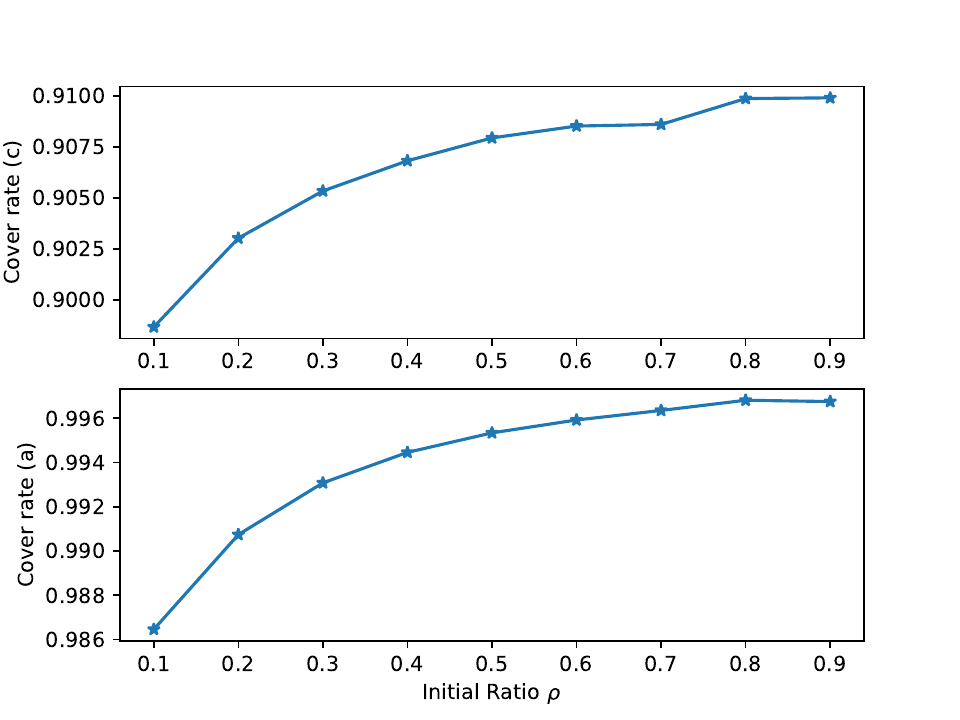}%
\label{fig:exp:filter conver rate}}
\hfil
\subfloat[FB15k]{\includegraphics[width=3.0in]{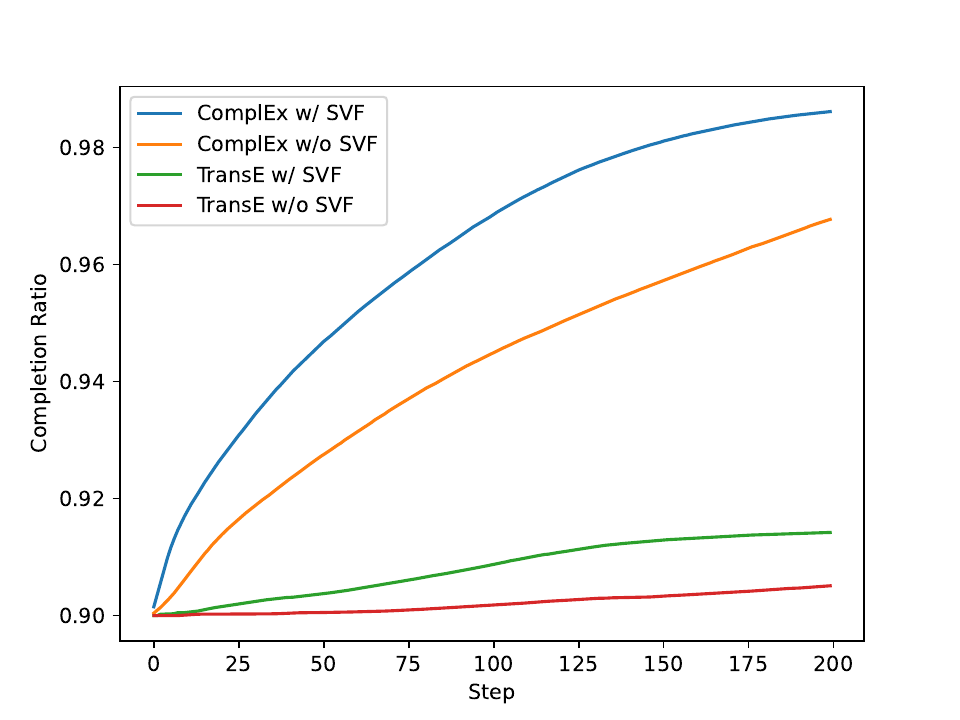}%
\label{fig:filter ablation}}
\caption{ (a) The cover rate of SVF in different initial ratio $\rho$. Since there are entities missing the class information, we consider two extreme situations. up: $c$ means conservative, which excludes all these entities. $a$ means actually, which includes these entities and is the case in real situation. (b) SVF ablation on performance.}
\label{fig:svf}
\end{figure*}

Initially, we establish a theoretical basis to affirm the feasibility of SVF. Figure \ref{fig:exp:filter conver rate} illustrates the coverage of SVF at various proportions represented by $\rho$. We examine two scenarios: the conservative case, where entities lacking class information are excluded from the coverage rate calculation, and the actual case, which includes these entities in the calculation. Both curves exhibit relatively minor variations across different values of $\rho$. In the actual case, the coverage rate is notably high, with an average of 0.99, a level of accuracy suitable for real-world applications. As an example, at $\rho = 0.8$, the data missed amounts to only $0.08\%$, a negligible fraction.

In preceding sections, we have established the effectiveness of SVF in terms of time efficiency. In this section, we assess SVF's effectiveness with regard to performance, focusing on the ComplEx and TransE models. Figure \ref{fig:filter ablation} demonstrates a significant enhancement in the models' performance resulting from the application of SVF. By using the ComplEx and TransE models for testing, we observe a significant improvement in both filtering efficiency and final completion rates following the integration of SVF. Clearly, without semantic filtering, numerous intrinsically unreasonable triples disrupt the model's filtering operations during each mining round, leading to reduced efficiency.

\begin{table}[!t]
    \centering
    \caption{MOAR of UniBi-O(2) and CP on low-resource WN18 and FB15k. }
    \resizebox{\linewidth}{!}{
    \begin{tabular}{lcccccc}
        \hline
        \multicolumn{1}{l}{} & \multicolumn{3}{c}{\textbf{WN18}}& \multicolumn{3}{c}{\textbf{FB15K}}\\
         Model & $\rho=0.3$ & $\rho=0.5$ & $\rho=0.7$ & $\rho=0.5$ & $\rho=0.7$ & $\rho=0.9$\\
         \hline
         CP & 0.189 & 0.343 & 0.515 & 0.428 & 0.510 & 0.588 \\
         RotatE & 0.307 & 0.364 & 0.768 & 0.513 & 0.610 & 0.621 \\
         RESCAL & 0.237 & 0.458 & 0.545 & 0.419 & 0.533 & 0.668\\
         UniBi-O(2) & 0.296 & 0.428 & 0.761 & 0.603 & 0.711 & 0.841 \\
         UniBi-O(3) & 0.275 & 0.451 & 0.762 & 0.630 & 0.720 & 0.846 \\
         \hline
    \end{tabular}}
    \label{tab:app:low resource moar}
\end{table}

\section{Supplementary Results of Low-resource PKGC}
\label{app:sec:supp_low_resource}

In addition to the CR@$k$ results presented in Table~\ref{tab:app:low resource cr}, we show the results for the MOAR metric in Table~\ref{tab:app:low resource moar}.

\end{document}